\documentclass{llncs}
\usepackage{amsmath}
\usepackage{graphicx}
\usepackage{float}
\usepackage{xcolor}
\usepackage{geometry}
\begin{document}

\title{forgeNet: a graph deep neural network model using tree-based ensemble classifiers for feature extraction}
\titlerunning{forest graph-embedded deep feedforward networks}  
%
\author{Yunchuan Kong\inst{1} \and Tianwei Yu\inst{1}}
\authorrunning{Kong and Yu} 
%
%
\institute{Department of Biostatistics and Bioinformatics, Emory University, 1518 Clifton Rd, Atlanta, GA 30322, USA\\
\email{tianwei.yu@emory.edu}
}

\maketitle              

\begin{abstract}
A unique challenge in predictive model building for omics data has been the small number of samples $(n)$ versus the large amount of features $(p)$. This ``$n\ll p$'' property brings difficulties for disease outcome classification using deep learning techniques. Sparse learning by incorporating external gene network information such as the graph-embedded deep feedforward network (GEDFN) \cite{kong2018graph} model has been a solution to this issue. However, such methods require an existing feature graph, and potential mis-specification of the feature graph can be harmful on classification and feature selection. To address this limitation and develop a robust classification model without relying on external knowledge, we propose a \underline{for}est \underline{g}raph-\underline{e}mbedded deep feedforward \underline{net}work (forgeNet) model, to integrate the GEDFN architecture with a forest feature graph extractor, so that the feature graph can be learned in a supervised manner and specifically constructed for a given prediction task. To validate the method's capability, we experimented the forgeNet model with both synthetic and real datasets. The resulting high classification accuracy suggests that the method is a valuable addition to sparse deep learning models for omics data.
\end{abstract}

\section{Introduction}
\label{Introduction}
In the study of bioinformatics, one important problem is the prediction of clinical outcomes using profiling datasets with a large amount of variables such as gene expression data. In such datasets, major challenges lie in the relatively small number of samples compared to the large number of predictors (genes), namely the ``$n\ll p$" issue. In addition, the complex unknown correlation structure among predictors results in more difficulty in prediction and feature selection. In order to tackle this challenging situation, machine learning approaches have been introduced for the prediction task \cite{cai2015classification,chen2014risk,kursa2014robustness,liang2013sparse,vanitha2015gene}. While the primary interest of these studies is to achieve high prediction accuracy, contributions have also been made for feature selection or learning effective feature representations \cite{cai2015classification,kursa2014robustness}. Based on the property of gene expression data, i.e., functionally associated genes tend to be statistically dependent and contribute to a biological outcome in a synergistic manner, a branch of classification research has been focused on integrating prior knowledge on the relations between genes into predictive models, in order to improve both classification performance and learning the structure of feature space. A critical data source to achieve this goal is the gene network constructed from existing biological knowledge, such as signal transduction network or protein-protein interaction network \cite{pmid25632107,pmid25859942}. A gene network is a graph-structured dataset with genes as the graph vertices and their functional relations as graph edges. In terms of classification tasks, each vertex in the gene network corresponds to a predictor, and it is expected that the gene network can provide useful information for a learning process. Motivated by this idea, certain classification methods have been developed where gene networks are integrated as additional information for the prediction and feature selection procedure. For example, support vector machines and traditional linear models such as logistic regression classifier can be modified by adding penalty terms to the objective function, where the penalty is defined according to pairwise distances between genes in a gene network \cite{kim2013network,lavi2012network,zhu2009network}. \cite{dutkowski2011protein} develops a random forest-based method, called Network-Guide Forest, where the feature sub-sampling in building decision trees is guided by graph search on the given gene network. Also, a recent study \cite{kong2018graph} brings gene networks to deep learning, where applications on omics data were restricted primarily due to the $n\ll p$ issue \cite{min2016deep}. In \cite{kong2018graph}, a deep learning model Graph-Embedded Deep Feedforward Network (GEDFN) is proposed with the gene network embedded as a hidden layer in deep neural networks to achieve an informative sparse structure. In GEDFN, the graph-embedded layer helps achieve two effects. One is model sparsity, and the other is the informative flow of information for prediction and feature evaluation. These two effects allow GEDFN to outperform other methods in gene expression classification given an appropriately specified feature graph.

Authors of these methods have demonstrated that combining gene networks with expression data results in better classification performance and more interpretable feature selection. However, these methods bear a common limitation, which is the potential mis-specification of the required gene network. In practice, gene expression data are used for various clinical outcomes, and the mechanistic relations between genes and different clinical outcomes can be quite different. Hence, there does not exist a known gene network that uniformly fits all classification problems. 
Thus, gene networks used in graph-embedded methods can only be ``useful" but not ``true". Consequently, how to decide if a known gene network is useful in predicting a certain clinical outcome with a certain gene expression dataset 
remains an unsolved problem, causing difficulties in applying graph-embedded methods in practice. \cite{kong2018graph} discusses the feature graph mis-specification issue of the GEDFN model and shows that the method is robust with mis-specified gene networks. Nevertheless, it is unrealistic to guarantee that the robustness applies in a broad sense, as feature graph structures can be extremely diverse such that simulation would not be able to cover all scenarios.

To address these issues, in this paper, we aim at developing a method that doesn't rely on a given feature network, yet can still benefit from the idea of building a model with sparse and informative flow of information. 
Instead of using known feature graphs, we try to construct a feature graph within the feature space. 
We propose a supervised feature graph construction framework using tree-based ensemble models, as literature shows that tree-based ensemble methods such as the Random Forest (RF) \cite{breiman2001random} and the Gradient Boosting Machine (GBM) \cite{friedman2002stochastic} are excellent tools for feature selection \cite{tang2014qualitative,vens2011random}. These tree-based methods also provide relational information between features in terms compensating each other in the classification task. 
We develop the \underline{for}est \underline{g}raph-\underline{e}mbedded deep feedforward \underline{net}work (forgeNet) model, with a built-in tree-based ensemble classifier as a feature graph extractor on top of a modified GEDFN model. The feature extractor selects features that span a reduced feature space, and constructs a graph between the selected features based on their directional relations in the decision tree ensemble. 

The application of tree-based ensemble methods as feature graph extractor is mainly based on two considerations: 1) the extractor selects effective features in a supervised manner. Thus the target outcome directly participates the feature graph construction. Compared to unsupervised feature construction such as using marginal or conditional correlation graphs, the resulting graph from trees is more informative and relevant to the specific classification task; 2) the feature extraction procedure helps reduce the dimension of the original feature space, alleviating the $n\ll p$ problem for the downstream neural network model. 

The paper is organized as follows: Section \ref{Methods} reviews the GEDFN model and illustrates our proposed forgeNet architecture. Section \ref{se} shows the simulation experiment results for comparing our new method with existing cutting-edge classifiers, followed by the real data analysis of a breast cancer dataset in Section \ref{rda}. Finally, a short conclusion is presented in Section \ref{Conclusion}. 

\section{Methods}
\label{Methods}
\subsection{Review of graph-embedded deep feedforward networks}
\label{gedfn}
We first briefly review the GEDFN model as our new method utilizes a similar neural network architecture. Recall a deep feedforward network with $l$ hidden layers:
\begin{align*}
Pr(\mathbf{y}|\mathbf{X},\mathbf{\Psi})&=softmax(\mathbf{Z}_{out}\mathbf{W}_{out}+\mathbf{b}_{out}) \\
	\mathbf{Z}_{out}&=\sigma(\mathbf{Z}_{l}\mathbf{W}_{l}+\mathbf{b}_l) \\
	\dots \\
	\mathbf{Z}_{k+1}&=\sigma(\mathbf{Z}_{k}\mathbf{W}_{k}+\mathbf{b}_k) \\
	\dots \\
	\mathbf{Z}_{1}&=\sigma(\mathbf{X}\mathbf{W}_{in}+\mathbf{b}_{in}), 
\end{align*}
where $\mathbf{X}\in \mathcal{R}^{n\times p}$ is the feature matrix with $n$ samples and $p$ features, $\mathbf{y}\in \mathcal{R}^n$ is the outcome containing classification labels, $\mathbf{\Psi}$ denotes all parameters, $\mathbf{Z}_{k}$ ($k=1,\dots,l-1,out$) are hidden layers with corresponding weights $\mathbf{W}_{k}$ and bias $\mathbf{b}_{k}$. The dimensions of $\mathbf{Z}$ and $\mathbf{W}$ depend on the number of hidden neurons $h_k$ ($k=1,\dots,l,in$) of each hidden layer, as well as the input dimension $p$ and the number of classes $h_{out}$. We mainly focus on binary classification problems hence the elements of $\mathbf{y}$ simply take binary values and $h_{out}\equiv 2$. The function $\sigma(\cdot)$ is the nonlinear activation such as sigmoid, hyperbolic tangent or rectifiers. The $softmax(\cdot)$ function converts values of the output layer into probability prediction. 

The graph-embedded feedforward net is a variant of the regular feedforward net with modified first hidden layer 
\begin{equation}
	\mathbf{Z}_{1}=\sigma(\mathbf{X}(\mathbf{W}_{in}\odot A)+\mathbf{b}_{in}) \label{g-layer}
\end{equation}
where $A$ is the adjacency matrix of a feature graph and $\odot$ is the Hadamard (element-wise) product. As in regular deep neural networks, the parameters to be estimated are all the weights and biases. The model is trained using a stochastic gradient decent (SGD) based algorithm by minimizing the cross-entropy loss function \cite{goodfellow2016deep}.  

\subsection{The forgeNet model}
\label{fgedfn}
Our newly proposed forest graph-embedded deep feedforward network (forgeNet) model consists of two components - the extractor component and the neural network component. The extractor component uses a forest model to select useful features from raw inputs with the supervision of training labels, as well as constructs a directed feature graph according to the splitting order in the individual decision trees. The neural network component feeds the generated feature graph and the raw inputs to GEDFN, and serves as the learner to predict outcomes. In forgeNet, a forest is defined as any ensemble of decision trees but not limited to random forests. In fact, any tree-based ensemble approach is applicable within the forgeNet framework. Besides RF and GBM mentioned in Section \ref{Introduction}, their variants with similar outputs are also possible options, or the forest can be simply built through bagging trees \cite{breiman1996bagging}. However, since RF and GBM models are the most commonly used tree ensembles, in this paper, we only employ these two methods for a proof-of-concept purpose. 

In forgeNet, a forest $\mathcal{F}$ is denoted as a collection of decision trees 
\begin{equation*}
	\mathcal{F}(\Theta) = \{\mathcal{T}_m(\Theta_m)\}, \ m=1,\dots, M,
\end{equation*}
where $M$ is the total number of trees in the forest, $\Theta=\{\Theta_1,\dots,\Theta_M\}$ represents the parameters, which include splitting variables and splitting values. 
In the feature graph extraction stage, $\mathcal{F}$ is fitted by training data $\mathbf{X}_{train}$ and training label $\mathbf{y}_{train}$, where $\mathbf{X}_{train}\in \mathcal{R}^{n_{train}\times p}$ and $\mathbf{y}_{train}\in \mathcal{R}^{n_{train}}$. After fitting the forest, we obtain $M$ decision trees, each of which contains a subset of features and their directed connections according to the tree splitting. At the same time, a binary tree can be viewed as a special case of a graph with directed edges. Hence, we can construct a set of graphs
\begin{equation*}
	\mathcal{G} = \{G_m(V_m, E_m)\}, \ m=1,\dots, M,
\end{equation*}
where $V_m$ and $E_m$ are collections of vertices and edges in $G_m$ respectively. Next, by merging all graphs in $\mathcal{G}$, the aggregated feature graph 
\begin{equation*}
	\mathbf{G}(V, E) = \bigcup_{m=1}^M G_m(V_m, E_m)
\end{equation*}
is obtained, where $V = \bigcup_{m=1}^M V_m$ and $E = \bigcup_{m=1}^M E_m$. 

In the form of its adjacency matrix, $\mathbf{G}$ is the feature graph to be embedded into the second stage of the forgeNet. Note that regardless which tree-based ensemble methods we use, it is likely that not all predictors in the original feature space can enter the forest model. A feature is included in $\mathbf{G}$ if and only if it is used at least once by the forest to split samples. As a result, the original feature space is reduced after the feature extraction. Denoting the number of vertices of $\mathbf{G}$ as $|V|$, we have $|V|<p$, and the input data matrix for the second stage is thus $\tilde{\mathbf{X}}_{train}\in \mathcal{R}^{n\times |V|}$. The columns in $\tilde{\mathbf{X}}_{train}$ corresponds to selected features in the original data $\mathbf{X}_{train}\in \mathcal{R}^{n\times p}$, and the order of columns does not matter. 

\begin{figure}
\centering
	\includegraphics[scale=0.4]{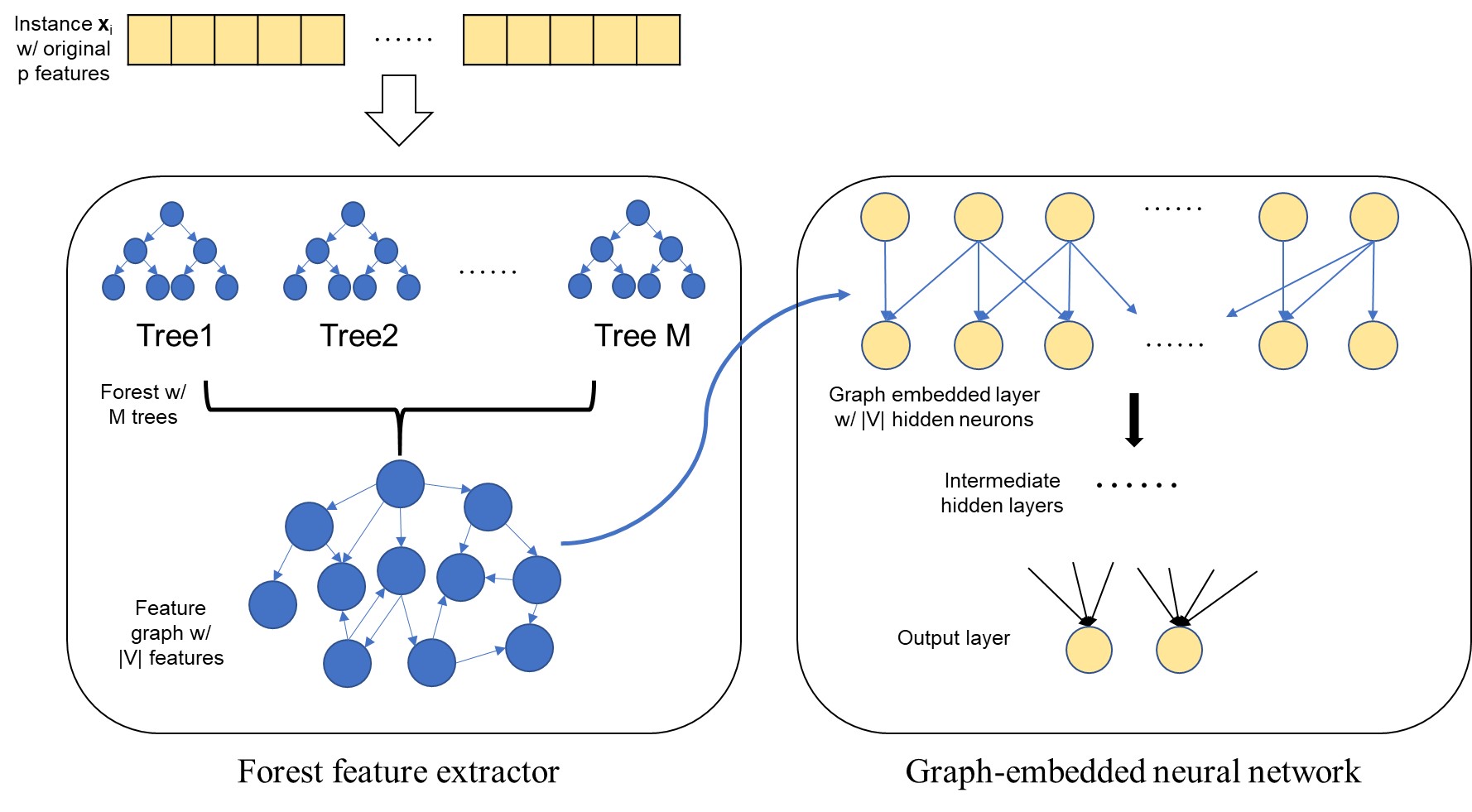}\\
\caption{Illustration of the forgeNet model. Notations are consistent with those in the text.}\label{architecture}
\end{figure}

The resulting feature graph $\mathbf{G}$ of feature extraction is a directed network, which differs from the one used in the original GEDFN. In \cite{kong2018graph}, the adjacency matrix $A$ in Eq. \ref{g-layer} represents an undirected feature graph. In the case of forgeNet, the adjacency matrix is naturally generalized to the directed version, and replacing $A$ in Eq. \ref{g-layer} with an asymmetric adjacency does not affect the model construction and training. A visualization of the entire forgeNet architecture is seen in Fig. \ref{architecture}. 

After fitting forgeNet with the training data, only the reduced input $\tilde{\mathbf{X}}_{test}$ and the testing label $\mathbf{y}_{test}$ are required for testing the prediction results, as $\tilde{\mathbf{X}}_{test}$ can be directly fed into the downstream neural nets together with the feature graph constructed from the forest. 

\subsection{Evaluation of feature importance}
The selection of predictors that significantly contribute to the prediction is another major aspect of gene expression data analysis, as they can reveal underlying biological mechanisms. Thus in forgeNet, we introduce a feature importance evaluation mechanism, which is closely related to the Graph Connection Weights (GCW) method proposed in \cite{kong2018graph} for the original GEDFN model. However, since the feature graph used in forgeNet has a different property from that in GEDFN where the feature graph is given, certain modifications of GCW are needed. 

The main idea of GCW is that, the contribution of a specific predictor is directly reflected by the magnitude of all the weights that are directly associated with the corresponding hidden neuron in the graph-embedded layer (the first hidden layer). In forgeNet, since the connection between the input layer and the first hidden layer is no longer symmetric due to the directed feature graph structure, to evaluate the importance of a given feature, we examine both hidden neurons in the first hidden layer and the nodes in the input layer. The importance score is thereby calculated as the summation of absolute values of the weights that are directly associated with the feature node itself and its corresponding hidden neuron in the graph-embedded layer:

\begin{equation*}\label{importance_score}
s_j=\sum_{u=1}^{p}|w_{ju}^{(in)}\mathcal{I}(A_{ju}=1)|+\sum_{v=1}^{p}|w_{vj}^{(in)}\mathcal{I}(A_{vj}=1)|+\sum_{m=1}^{h_1}|w_{jm}^{(1)}|,\quad j=1,\dots,p,
\end{equation*}

where $s_j$ is the importance score for feature $j$, $w^{(in)}$ denotes weights between the input and first hidden layers, and $w^{(1)}$ denotes weights between the first hidden layer and the second hidden layer. The score consists of three parts: the first two terms summarize the importance of a feature according to the directed edge connection in the feature graph $\mathbf{G}$; the third term summarizes the contribution of the feature according to the connection with the second hidden layer $\mathbf{Z}_{2}$. Note that the input data $\mathbf{X}$ are required to be Z-score transformed (the original value minus the mean across all samples and then divided by the standard deviation), ensuring all variables are of the same scale so that the magnitude of weights are comparable. Once the forgeNet is trained, the importance scores for all the variables can be calculated using trained weights. 



\section{Implementation}
\label{app}
The method is available in Python at \url{https://github.com/yunchuankong/forgeNet}. We employ the Scikit-learn \cite{scikit-learn} package for the implementation of RF, the Xgboost package \cite{Chen:2016:XST:2939672.2939785} for GBM, and the Tensorflow library \cite{tensorflow2015-whitepaper} for deep neural networks. For the choice of activation functions of neural nets, the rectified linear unit (ReLU) \cite{nair2010rectified} is employed. This non-linear activation has an advantage over the sigmoid function and the hyperbolic tangent function as it avoids the vanishing gradient problem \cite{hochreiter2001gradient} during model training. The entire neural net part of forgeNets is trained using the Adam optimizer \cite{DBLP:journals/corr/KingmaB14}, which is the state-of-the-art version of the popular stochastic gradient descent algorithm. Also, we use the mini-batch training strategy by which the optimizer loops over randomly divided small proportions of the training samples in each iteration. Details about the Adam optimizer and the mini-batch strategy applications in deep learning can be found in \cite{goodfellow2016deep,DBLP:journals/corr/KingmaB14}.

The performance of a deep neural network model is associated with many hyper-parameters, including the number of hidden layers, the number of hidden neurons in each layer, the dropout proportion of training, the learning rate and the batch size. As the hyper-parameters are not of primary interest in our research, in the simulation and real data experiments, we simply tune hyper-parameters using grid search in a feasible parameter space. Also, since our experiments contains a number of datasets, it is not plausible to fine tune models for each dataset. Instead, we tune hyper-parameters using some preliminary synthetic datasets, and apply the set of parameters to all experimental data. For simulation experiments, the number of trees of our forgeNets is 1000 and the number of hidden layers of the neural net is three with $p$ (graph-embedded layer), 64 and 16 hidden neurons respectively. For real data analyses, we have 2500 trees in the forest part since the feature space is much larger, and the neural net structure is the same as it is in simulation.  

\section{Simulation experiments}
\label{se}
The goal of the simulation experiments is to mimic disease outcome classification using gene expression data with $n\ll p$. Effective features are sparse and potentially correlated through an underlying unknown structure. Several benchmark methods are experimented in addition to the new forgeNet model for comparison purpose. Through simulation, we intend to investigate whether the forgeNet model is able to outperform other classifiers without knowing the underlying structure of features. 

\subsection{Synthetic data generation}
\label{sdg}
We follow a similar procedure described in \cite{kong2018graph}. For a given number of features $p$, the preferential attachment algorithm (BA model) \cite{barabasi1999emergence} is employed to generate a scale-free network as the underlying true feature graph. Defining the distance between two features in the network as the shortest path between them, we calculate the $p\times p$ matrix $D$ recording pairwise distances among features. Next, the distance matrix is transformed into a covariance matrix $\Sigma$ by letting
\begin{equation*}
	\Sigma_{ij}=0.6^{D_{ij}}, i,j=1,\dots,p.
\end{equation*} 

After obtaining the covariance matrix between features, we generate $n$ multivariate Normal samples as the data matrix $\mathbf{X}=(\mathbf{x}_1,\dots,\mathbf{x}_n)^T$ i.e.
\begin{equation*}
	\mathbf{x}_i\sim \mathcal{N}(\mathbf{0},\Sigma), i=1,\dots\,n,
\end{equation*}
where $n\ll p$ for imitating gene expression data. To generate outcome variables, we first select a subset of features to be ``true" predictors. Among vertices with relatively high degrees (``hub nodes") in the feature graph, part of them are randomly selected as ``cores", and a proportion of the neighboring vertices of cores are also selected. Denoting the number of true predictors as $p_0$, we uniformly sample a set of parameters $\mathbf{\beta}=(\beta _1,\dots,\beta _{p_0})^T$ and an intercept $\beta_0$ from a small range, say $(-0.15, 0.15)$. 
Finally, the outcome variable $\mathbf{y}$ is generated through a procedure similar to the generalized linear model framework
\begin{equation*}
    	y_i=\mathcal{I}\{g(\beta_0 + (\mathbf{x_i}^{(true)})^T\mathbf{\beta})>t\},\quad i=1,\dots\,n,
\end{equation*}
where $\mathbf{x_i}^{(true)}\in \mathcal{R}^{p_0}$ is the sub-vector of $\mathbf{x_i}$ and $t$ is a threshold. For the transformation function $g(\cdot)$, we consider a weighted sum of hyperbolic tangent and quadratic function  
\begin{equation*}
	g(x) = 0.7\phi(tanh(x))+0.3\phi(x^2).
\end{equation*} 
The reason of using this $g(\cdot)$ function is that the transformation is non-monotone, which brings in more challenges for classification. The function $\phi(\cdot)$ is the min-max transformation scaling the input to $[0,1]$, i.e., the original value minus the sample minimum and then divided by the difference between the sample maximum and the sample minimum.  

Following the above data generation scheme, we simulate a set of synthetic datasets with $p=5,000$ features and $n=400$ samples. Since in gene expression data, the true signals for a certain prediction task are sparse ($p_0\ll p$), We choose $p_0=15, 30, 45, 60$ and $75$ as the numbers of true predictors, corresponding to $1$ to $5$ cores selected among all hub nodes in the feature graph. 

\subsection{Evaluation of simulation experiments}

We compare our method with several benchmark models. First, since the true feature graphs are known for simulation data, we are able to test the original GEDFN model with correctly specified feature graphs. At the same time, we also experiment GEDFN with mis-specified feature graphs by randomly generating Erdo-Renyi random graphs \cite{erdos1959random}, which have a different graph topology structure from the true scale-free networks. Also, since forgeNet inherently fits a tree-based ensemble classifier, it is natural to compare the performance of a forgeNet with its forest part alone. We choose two representative tree methods RF and GBM for the experiments, and correspondingly test two versions of forgeNets - forgeNet(RF) and forgeNet(GBM). Finally, the logistic regression classifier with lasso (LRL) \cite{tibshirani1996regression} is also added as a representative of linear machines. 
   
For each of the data generation settings, ten independent datasets are generated. For each dataset, we randomly split samples into training and testing sets at a ratio of 4:1. All models are fitted using the training dataset and then used to predict the testing dataset. To evaluate classification results, areas under Receiver Operating Characteristic curves (ROC-AUC) are calculated using the predicted class probabilities and the labels of the testing set. The final testing result for a simulation case is then given by the average testing ROC-AUC across the ten datasets.

As for feature selection, all the methods except LRL provide relative feature importance scores; LRL does not rank features but directly gave the selected feature subset. Knowing the true predictors for simulated data, we could use the binary true predictor labels to evaluate the accuracy of feature selection. However, in preliminary numerical experiments, it is observed that though we fix the number of true features in each case, neighboring features of true predictors in the feature graph are also informative for classification even if they are not in the true feature set. This is because these neighboring features have a relatively high correlation with selected true predictors ($0.6$ according to Section \ref{sdg}). Therefore, when evaluating the results of feature selection, it is more appropriate to investigate a set of ``relevant" features including those neighboring features, rather than the ``true" feature set only. The average numbers of relevant features are 208.8, 460.4, 615.4, 717.8, and 864.7 respectively, corresponding to the five cases of true features $p_0=15, 30, 45, 60$ and $75$. 

Since the relevant feature sets are still small compared to the entire feature space ($p=5000$), the AUC of the precision-recall curve is a more appropriate metric here. We thus compare feature selection results using binary labels of relevant features for all methods providing feature scores. As for LRL, for each dataset, we compare recall values of our methods and LRL given the precision value of LRL. That is, the precision of LRL helps locate points on the precision-recall curves of forgeNets, and corresponding recall values are used for comparison.

\subsection{Simulation results}
\label{sr}

\begin{figure}
\centering
	\begin{minipage}[b]{0.32\textwidth}
		\includegraphics[width=\textwidth]{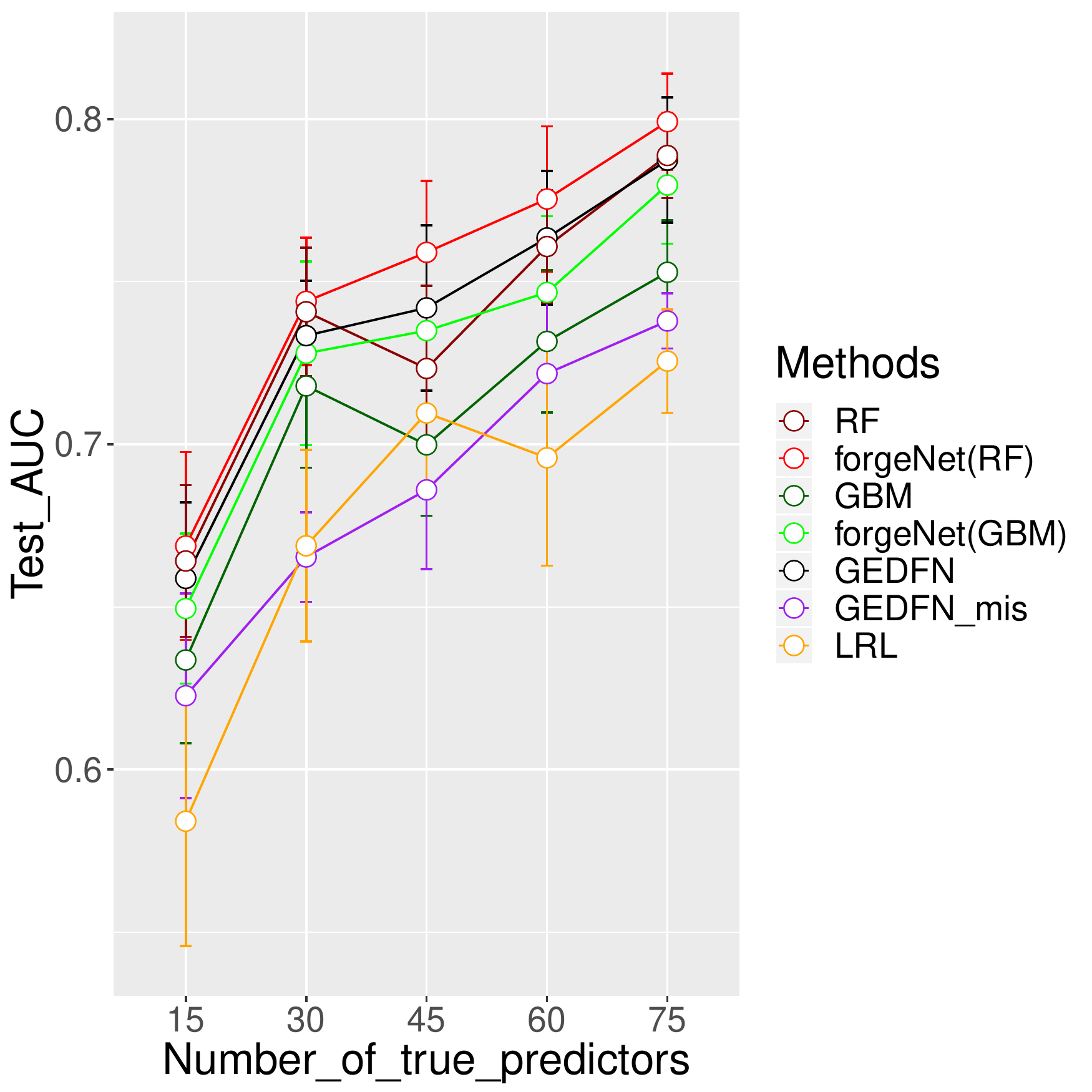}\\
        \centering{(a)}
	\end{minipage}
    \begin{minipage}[b]{0.32\textwidth}
		\includegraphics[width=\textwidth]{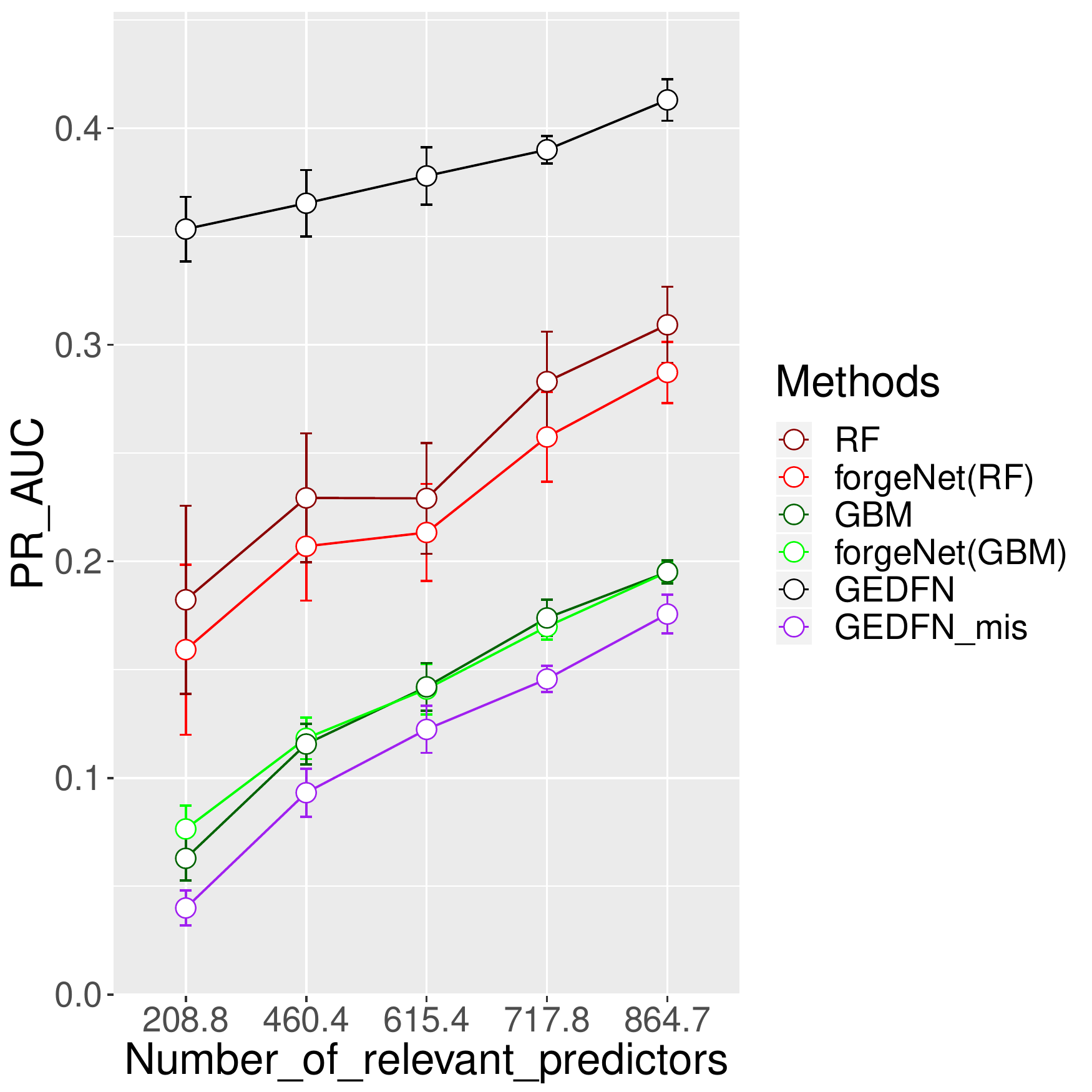}\\
        \centering{(b)}
	\end{minipage}
    \begin{minipage}[b]{0.32\textwidth}
		\includegraphics[width=\textwidth]{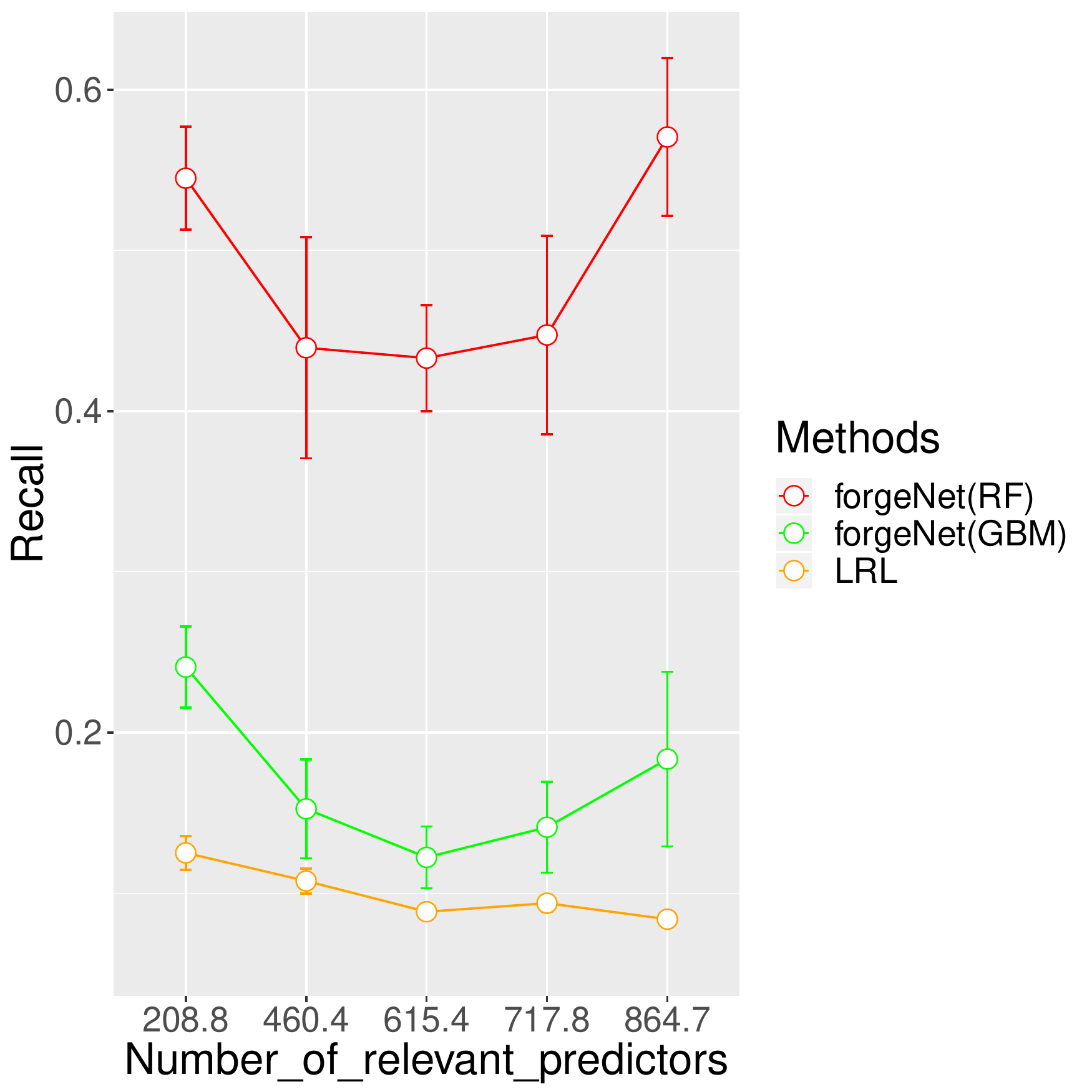}\\
        \centering{(c)}
	\end{minipage}
\caption{Comparison of classification and feature selection for the simulation study. (a) AUC of ROC for classification; (b) AUC of precision-recall for feature selection; (c) recall plots given fixed precision from LRL. Error bars represent the estimated mean quantities plus/minus the estimated standard errors.}\label{simulation_figures}
\end{figure}

Fig. \ref{simulation_figures}(a) shows the results of classification accuracy comparison. With the increasing number of true predictors, all of the methods performed better as there were more signals in the entire feature space. From the figure, the two versions of forgeNets, forgeNet(RF) and forgeNet(GBM), significantly improved the classification performance of their forest counterparts, i.e., RF and GBM. Also, the forgeNets achieved similar classification accuracy as GEDFN which benefited from the use of true feature graphs, and forgeNet(RF) was the only method that outperformed GEDFN. When GEDFN was given mis-specified feature graphs (GEDFN\_mis), its classification ability was weakened with AUC values similar to LRL. In summary, in terms of prediction, forgeNets beat all classic machine learning methods compared here (RF, GBM, LRL), achieved similar or even better accuracy compared to GEDFN using true feature graphs, and significantly outperformed GEDFN once its feature graphs were mis-specified. 

Feature selection results can be seen in Fig. \ref{simulation_figures}(b) and (c). Comparing the precision-recall AUCs from Fig. \ref{simulation_figures}(b), it can be observed that GEDFN using true feature graph was the best method for feature importance ranking, yet again the outstanding performance was ruined by mis-specified feature graphs. The results of forgeNets were significantly better than GEDFN\_mis, and were consistent with their forest counterparts. As the training of neural networks in forgeNets largely relied on feature graphs given by forests, it is not surprising to see that forgeNets could achieve similar feature selection results as their forest counterparts. In Fig. \ref{simulation_figures}(c), both forgeNet(RF) and forgeNet(GBM) were able to achieve higher recall values than LRL. In summary, in terms of feature selection, forgeNets outperformed the traditional lasso method and had consistent performance with their forest counterparts. Although not as good as GEDFN with true feature graphs, forgeNets produced significantly better feature selection than GEDFN using mis-specified feature graphs. Finally, we observe that the choice of the forest in forgeNets mattered, and among the two versions in our experiments, forgeNet(RF) was a more powerful model.         

The simulation study proved the forgeNet a powerful classifier, with reasonably good feature selection ability. Through the experiment results, one can easily conclude the novelty of forgeNets is that, by borrowing the neural net architecture of the original GEDFN, forgeNets utilize feature information more effectively in classification tasks compared to regular tree-based ensemble methods.

\section{Real data applications}
\label{rda}
\subsection{Datasets}

We applied forgeNets to the Cancer Genome Atlas (TCGA) breast cancer (BRCA) RNA-seq dataset \cite{koboldt2012comprehensive}. The dataset consists of a gene expression matrix with 20155 genes and 1097 cancer patients, as well as the clinical data including survival information. The classification task is to predict the three-year survival outcome. We excluded patients with missing or censored survival time for which the three-year survival outcome could not be decided. Also, genes with more than 10\% of zero values were also screened out. As a result, the final dataset contains a total of $p=16027$ genes and $n=506$ patients, with 86\% positive cases. For each gene, its expression value was Z-score transformed. 

Using the BRCA data, we again tested two versions of forgeNets together with RF, GBM, and LRL. 
The classification was conducted using a 5-fold stratified cross validation process, and the final prediction AUC for each method is computed by averaging the five validation results. 

\subsection{Results}
Table \ref{BRCA_res} summarizes the classification results. From the table, forgeNets again outperformed their forest counterpart models and LRL. Therefore, the real data application also led to a similar conclusion as in Section \ref{se} that forgeNets brought in significant improvement for classification.

\begin{table}
\centering
\normalsize 
\caption{Classification results for BRCA data}
\begin{tabular}{c|ccccc}
\hline 
Methods & forgeNet(RF) & RF & forgeNet(XGB) & XGB & LRL\tabularnewline
\hline 
Mean AUC  & 0.742  & 0.672  & 0.716 & 0.691 & 0.689 \tabularnewline
\hline 
s.d.  & 0.066 & 0.048 & 0.100 & 0.022 & 0.084\tabularnewline
\hline 
\end{tabular}
\label{BRCA_res}
\end{table}

Feature selection was also conducted for BRCA data. We obtained ranked gene importance lists by averaging importance scores across the five cross validation results from all methods except LRL. For LRL, the intersection (456 genes) of the five selected feature sets is used as the final selected features. We chose top 500 ranked genes for each ranked list so that the numbers are of a similar magnitude as the genes selected by LRL. Functional analysis of all final gene lists was conducted by the Gene Ontology (GO) enrichment test using GOstats package \cite{pmid17098774}. We limited the analysis to GO biological processes containing 10-500 genes, and a p-value cutoff of 0.005. After manual removal of highly overlapping GO terms, the top 3 GO terms that contained the most number of selected genes are found in Table \ref{GO_BRCA}.

\begin{table}[H]
\centering
\small 
\caption{Top 3 GO biological processes for each method, after manual removal of redundant GO terms.}\label{GO_BRCA}

\begin{tabular}{lllll}
\hline 
\textbf{ID} & \textbf{Term} & \textbf{P\_value} & \textbf{Count} & \textbf{Size}\tabularnewline
\hline 
\hline 
forgeNet(RF) &  &  &  & \tabularnewline
\hline 
GO:0031647 & regulation of protein stability & 0.00123 & 17 & 229\tabularnewline
GO:0090502 & RNA phosphodiester bond hydrolysis, endonucleolytic & 0.00369 & 7 & 62\tabularnewline
GO:1901998 & toxin transport & 0.00499 & 5 & 35\tabularnewline
\hline 
RF &  &  &  & \tabularnewline
\hline 
GO:2000679 & positive regulation of transcription regulatory region DNA binding & 0.00255 & 4 & 19\tabularnewline
GO:0010172 & embryonic body morphogenesis & 0.00313 & 3 & 10\tabularnewline
GO:0090042 & tubulin deacetylation & 0.0042 & 3 & 11\tabularnewline
\hline 
forgeNet(GBM) &  &  &  & \tabularnewline
\hline 
GO:0001676 & long-chain fatty acid metabolic process & 0.00138 & 9 & 84\tabularnewline
GO:0032890 & regulation of organic acid transport & 0.00155 & 6 & 40\tabularnewline
GO:0046470 & phosphatidylcholine metabolic process & 0.00449 & 7 & 65\tabularnewline
\hline 
GBM &  &  &  & \tabularnewline
\hline 
GO:0006633 & fatty acid biosynthetic process & 0.000454 & 12 & 121\tabularnewline
GO:0030520 & intracellular estrogen receptor signaling pathway & 0.000643 & 7 & 47\tabularnewline
GO:0010763 & positive regulation of fibroblast migration & 0.00322 & 3 & 10\tabularnewline
\hline 
LRL &  &  &  & \tabularnewline
\hline 
GO:0051047 & positive regulation of secretion & 0.000609 & 20 & 317\tabularnewline
GO:0006090 & pyruvate metabolic process & 0.000911 & 9 & 90\tabularnewline
GO:0019359 & nicotinamide nucleotide biosynthetic process & 0.00204 & 8 & 82\tabularnewline
\hline
\end{tabular}

\end{table}

The top GO term selected by forgeNet(RF) was regulation of protein stability. It has been found that estrogen receptor (ER) alpha has increased abundance and activity in breast cancer. One of the mechanisms facilitating this change is the protection of ER from degradation by the ubiquitin-proteasome system \cite{pmid27561704}. Another critical protein, HER2 (human epidermal growth factor receptor 2), has also been found to have increased stability and activity in some breast cancer tissues through the formation of Her2-Heat-shock protein 27 (HSP27) complex \cite{pmid18834540}. The protein stability mechanism has not been previously linked to the survival outcome of breast cancer. The second GO term found by forgeNet(RF), RNA phosphodiester bond hydrolysis, endonucleolytic, is part of rNRA and tRNA processing. It plays a critical role in the protein synthesis of the cancer cells. The third term, toxin transport, is specific to breast cancer. It is suggested that increased toxin presence in the mammary tissue is a pre-disposing factor to breast cancer \cite{pmid12706546,Quezada_2014}. 

The forgeNet(GBM) and GBM results both point to fatty acid metabolism, which is known to be dysregulated in breast cancer \cite{pmid28412757}. The GBM selected the estrogen receptor signaling pathway, which is critically important in breast cancer development. The LRL selected GO terms include positive regulation of secretion, which includes lactation, in addtion to metabolic processes. 

\section{Conclusion}
\label{Conclusion}
We presented forgeNet that uses tree-based ensemble methods to extract feature connectivity information, and uses GEDFN for graph-based predictive model building. The new method was able to achieve sparse connection for neural nets without seeking external information, i.e., known feature graphs. It works well in the ``$n\ll p$" situation. Simulation experiments showed forgeNets' relatively higher classification accuracy compared to existing methods, and a TCGA RNA-seq dataset demonstrated the utility of forgeNets in both classification and the selection of biologically interpretable predictors. 

\section*{Acknowledgements}
\label{Acknowledgements}
This work was partially supported by NIH grant R01GM124061.

\bibliographystyle{splncs03}
\bibliography{references} 

\end{document}